\relax
\documentclass[letterpaper]{article} 
\usepackage{aaai22}  
\usepackage{times}  
\usepackage{helvet}  
\usepackage{courier}  
\usepackage[hyphens]{url}  
\usepackage{graphicx} 
\usepackage{natbib}  
\usepackage{caption} 
\DeclareCaptionStyle{ruled}{labelfont=normalfont,labelsep=colon,strut=off} 
\usepackage{algorithm}
\usepackage{algorithmic}
\usepackage{amsfonts}
\usepackage{amsbsy}
\usepackage{amsmath}
\usepackage{multirow}
\usepackage{booktabs}
\usepackage{newfloat}
\usepackage{listings}
\floatstyle{ruled}
\newfloat{listing}{tb}{lst}{}
\floatname{listing}{Listing}
\usepackage{color} 

\title{Feature Generation and Hypothesis Verification for Reliable Face Anti-Spoofing}
\author{
    Shice Liu\textsuperscript{\rm 1}\equalcontrib,
    Shitao Lu\textsuperscript{\rm 1,2}\equalcontrib,
    Hongyi Xu\textsuperscript{\rm 1,3},
    Jing Yang\textsuperscript{\rm 1}\thanks{These authors are the corresponding authors.},
    Shouhong Ding\textsuperscript{\rm 1}$^{\dag}$,
    Lizhuang Ma\textsuperscript{\rm 2,3}
}
\affiliations{
    \textsuperscript{\rm 1}Youtu Lab, Tencent, Shanghai, China\\
    \textsuperscript{\rm 2}East China Normal University, Shanghai, China\\
    \textsuperscript{\rm 3}Shanghai Jiao Tong University, Shanghai, China\\
    \{shiceliu,alonyang,ericshding\}@tencent.com; lusto32768@gmail.com; xuhongyi@sjtu.edu.cn; lzma@cs.ecnu.edu.cn
}

\begin{document}

\maketitle

\begin{abstract}
Although existing face anti-spoofing (FAS) methods achieve high accuracy in intra-domain experiments, their effects drop severely in cross-domain scenarios because of poor generalization. Recently, multifarious techniques have been explored, such as domain generalization and representation disentanglement. However, the improvement is still limited by two issues: 1) It is difficult to perfectly map all faces to a shared feature space. If faces from unknown domains are not mapped to the known region in the shared feature space, accidentally inaccurate predictions will be obtained. 2) It is hard to completely consider various spoof traces for disentanglement. In this paper, we propose a Feature Generation and Hypothesis Verification framework to alleviate the two issues. Above all, feature generation networks which generate hypotheses of real faces and known attacks are introduced for the first time in the FAS task. Subsequently, two hypothesis verification modules are applied to judge whether the input face comes from the real-face space and the real-face distribution respectively. Furthermore, some analyses of the relationship between our framework and Bayesian uncertainty estimation are given, which provides theoretical support for reliable defense in unknown domains. Experimental results show our framework achieves promising results and outperforms the state-of-the-art approaches on extensive public datasets.

\end{abstract}

\section{Introduction}

Nowadays, face recognition (FR) has been widely used in many AI systems in our daily lives.
However, endless face presentation attacks 
continuously threaten the security of face applications. To endow the AI systems with this
important defensive capability, FAS techniques must be equipped.

In the past few years, various FAS approaches have been proposed. Some representative methods are handcrafted feature-based \cite{LBP00}, movement-based \cite{pan2007eyeblink}, distortion-based \cite{wen2015face}, physiological signal-based \cite{li2016generalized} and deep feature-based \cite{yang2014learn}. Although these methods perform well in intra-domain experiments, the effects decrease severely in cross-domain scenarios due to poor generalization. With the aim of more generalized models, two categories of methods are currently being studied.

Domain generalization-based methods are exploited to learn a domain-agnostic feature space. Intuitively, if a model works well in several known domains, it would be more likely to be effective in other unknown domains. To achieve it, commonly used solutions include metric learning \cite{jia2020single}, adversarial training \cite{liu2021dual}, meta-learning \cite{wang2021self} and special structures \cite{yu2021dual}. Despite using different techniques, all these methods aim to attain a better feature extraction backbone so that regardless of which domain the input comes from, the backbone always outputs generalized features. However, in real scenarios, it is hard to perfectly map real and fake faces from different domains to a generalized shared feature space. Additionally, from the perspective of Bayesian uncertainty estimation, models will be prone to giving wrong results if they are fed with what they do not know.

\begin{figure}[t!]
    \centering
    \includegraphics[width = \linewidth]{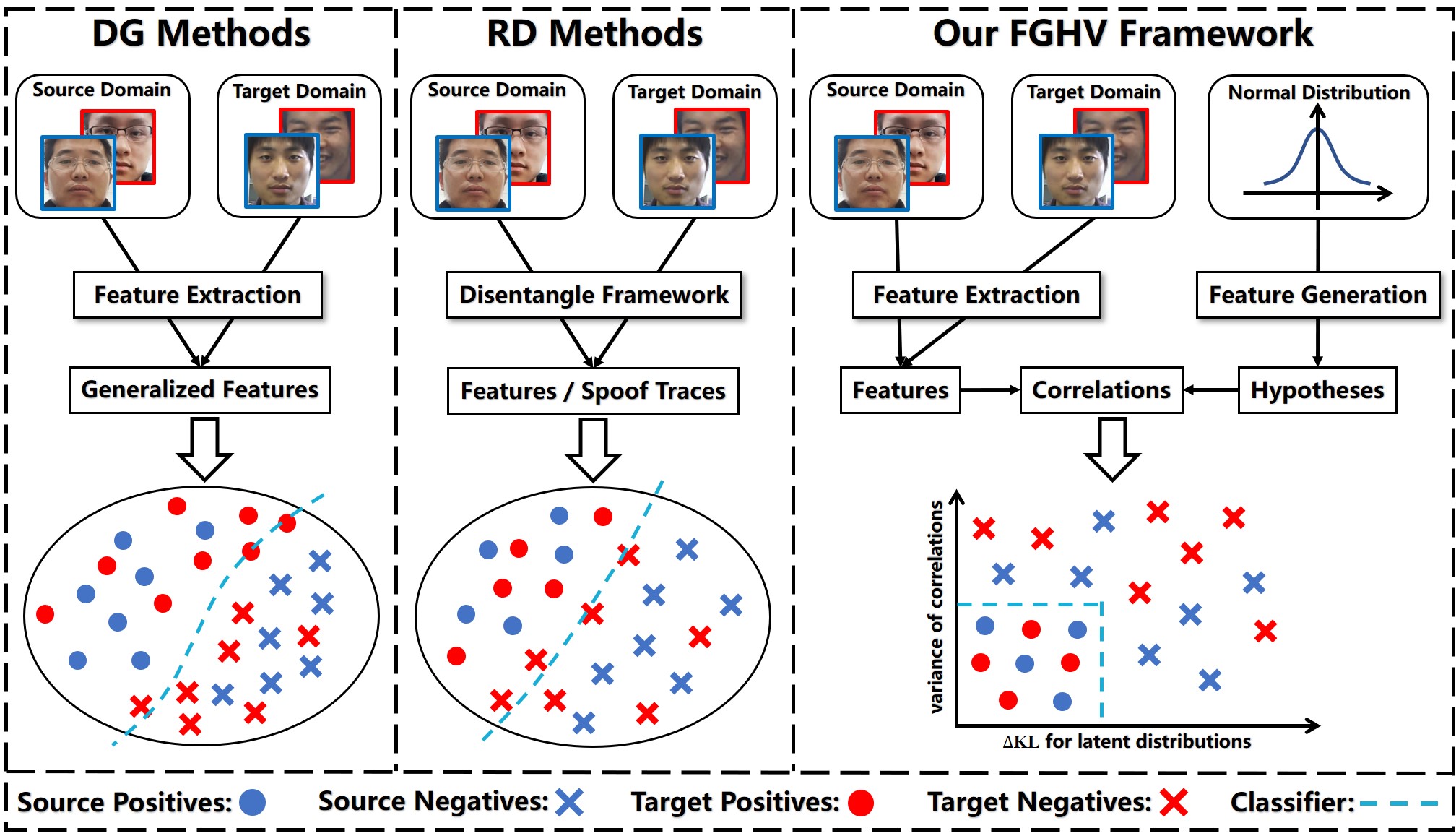}
    \caption{Comparison among domain generalization-based methods (DG), representation disentanglement-based methods (RD) and our FGHV framework.}
    \label{fig:compare_with_others}
\end{figure}

Representation disentanglement-based methods hold the view that features can be partitioned into liveness-related parts (i.e., spoof trace) and liveness-unrelated parts (e.g., appearance, identification, age), and only the liveness-related parts are leveraged to classify for better generalization. To this end, CycleGAN \cite{zhang2020face} or arithmetic operations on images \cite{liu2020disentangling} is utilized to extract the spoof trace. However, spoof traces vary with the type and style of attacks. That's to say, these methods might be confused if inputs from unknown domains are given.

In order to tackle such issues, we propose a Feature Generation and Hypothesis Verification (FGHV) framework to verify 
both real-face feature space and real-face distribution by generating hypotheses. Fig. \ref{fig:compare_with_others} depicts the comparison to two related methods. Firstly, we leverage two feature generation networks to generate features of real faces and known attacks respectively which are also termed as hypotheses. Different from domain generalization-based methods that construct a shared feature space from raw images in huge RGB space, hypotheses are generated from latent vectors which are sampled from the same distribution during training and testing periods. Compared with representation disentanglement-based methods that mine spoof traces, we mainly care about features of real faces which are more similar across domains. Secondly, we devise a Feature Hypothesis Verification Module (FHVM) to estimate to what extent the input face comes from the real-face feature space. Specifically, after generating enough real-face hypotheses via the feature generation network, the FHVM evaluates the consistency of correlations between each hypothesis and the input face. Thirdly, we design a Gaussian Hypothesis Verification Module (GHVM) to measure the KL divergence between the input face distribution and the real-face distribution in the latent space. Furthermore, from the viewpoint of Bayesian uncertainty estimation, we analyze that our framework actually constructs a more effective prior distribution than Bayesian neural network \cite{shridhar2018uncertainty, shridhar2018bayesian,farquhar2020radial} to estimate the epistemic uncertainty, which brings greater effects and better reliability.

The main contributions are summarized as follows:

$\bullet$ The FAS problem is modeled as a classification problem of real faces and non-real faces, and to the best of our knowledge, feature generation networks for producing hypotheses are introduced into the FAS task for the first time.

$\bullet$ Two effective hypothesis verification modules are proposed to judge whether the input face comes from the real-face feature space and the real-face distribution respectively.


$\bullet$ The relationship between our framework and Bayesian uncertainty estimation is clearly stated. And comprehensive experiments and visualizations demonstrate the effectiveness and reliability of our approach.

\section{Related Work}

We briefly review related works on traditional FAS methods, and then detail domain generalization-based methods and representation disentanglement-based methods which are two popular research orientations based on deep learning. Finally, we give a sketch of Bayesian deep learning.

\textbf{Traditional Face Anti-Spoofing.} Early researchers have introduced lots of handcrafted features to achieve FAS task, such as LBP \cite{LBP00,LBP01,LBP02,LBP03}, HOG \cite{HoG00,HoG01} and SIFT \cite{SIFT00}. Since they are too simple to perform well, more liveness cues are explored later, such as eye blinking \cite{pan2007eyeblink}, face movement \cite{wang2009face}, light changing \cite{zhang2021aurora} and remote physiological signals (e.g., rPPG \cite{li2016generalized,liu2018learning,yu2021transrppg,hu2021end}). However, these methods are always limited by low accuracy or complicated process in video data.

\textbf{Domain Generalization-Based Face Anti-Spoofing.} Although deep learning facilitates the FAS task, the generalization ability for multiple domains still need to be improved. To this end, researchers have tapped the potential of various techniques. Some approaches measured and constrained the distance of features or domains to obtain domain-agnostic features. For instance, \citet{li2018unsupervised} used the MMD distance to make features unrelated with domains. \citet{jia2020single} and \citet{yang2021few} introduced triplet loss \cite{li2019group,li2019detecting} and \citet{zhang2021structure} even constructed a similarity matrix to constrain the distance between features. Also, many meta-learning-based methods were exploited to find a generalized space among multiple domains \cite{shao2020regularized, qin2020learning, kim2021domain, chen2021generalizable, wang2021self,qin2021meta}. Besides, \citet{wang2019improving} and \citet{liu2021dual} utilized adversarial training, while \citet{yu2020face,yu2020searching,yu2020auto,yu2020multi,yu2021dual,yu2021revisiting} and \citet{chen2021dual} proposed some special network structures and loss functions for better generalization. Although the effects were improved from different aspects, all these methods intended to make the feature extraction backbone generalized. Nevertheless, mapping all real faces and attacks of different domains to a shared feature space is difficult and the shared feature space is usually not generalized well. Furthermore, in the view of uncertainty estimation, given some inputs from unknown domains, these models might produce surprisingly bad results because of incapacity to distinguish what they know and what they don't.

\textbf{Representation Disentanglement-Based Face Anti-Spoofing.} Some other representative methods for generalization realized FAS through detecting spoof traces which were disentangled from input faces. \citet{stehouwer2020noise} would like to synthesize and identify noise patterns from seen and unseen medium/sensor combinations, and then benefited from adversarial training. \citet{liu2020disentangling} intended to disentangle spoof traces via arithmetic operations, adversarial training and so on. Especially, motivated by CycleGAN \cite{zhu2017unpaired}, \citet{zhang2020face} finished disentangling liveness features of real faces and attacks by exchanging, reclassification and adversarial training. Generally speaking, these methods successfully disentangled liveness-related features and thus had relatively stronger interpretability. 
However, the set of attacks is an open set and it is almost impossible to construct the distribution of attacks or disentangle diverse spoof traces accurately.

\begin{figure*}[t!]
    \centering
    \includegraphics[width = \linewidth]{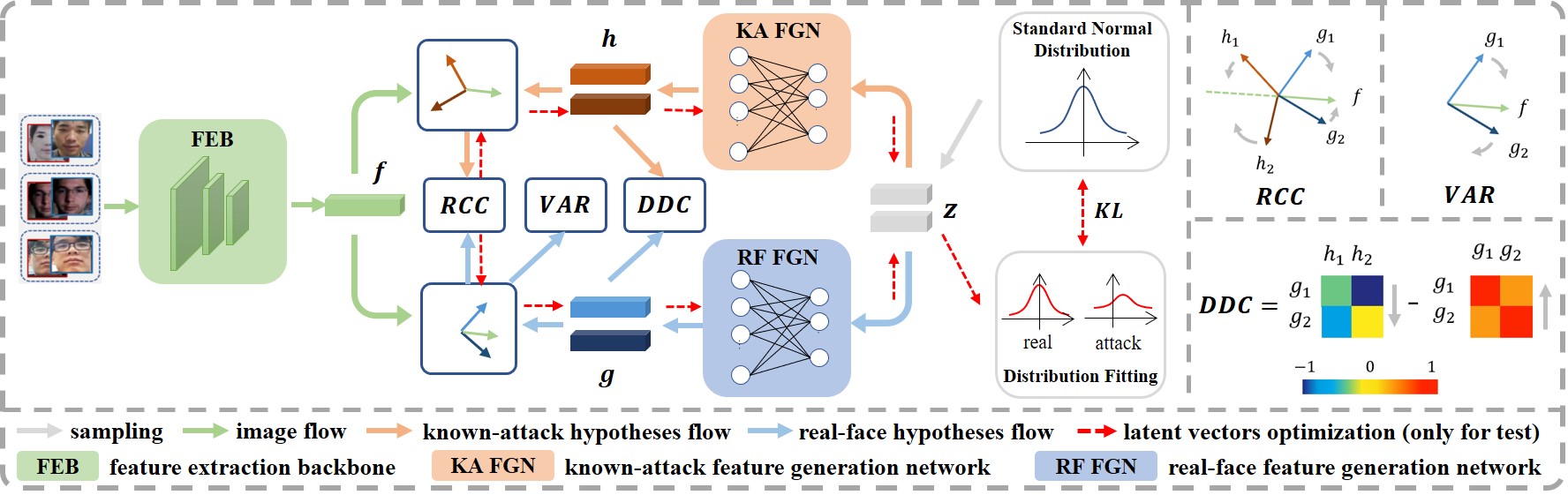}
    \caption{An overview of the proposed \textbf{Feature Generation and Hypothesis Verification} framework. \textbf{Feature Hypothesis Verification Module} introduces the variance constraint (VAR) in both training and testing. Besides, \textbf{Gaussian Hypothesis Verification Module} draws support from Relative Correlation Constraint (RCC) and Distribution Discrimination Constraint (DDC) in training, and then acquires the distribution distance (i.e., KL divergence) via latent vectors optimization in testing. The gray arrows on the right prompt how hypotheses are optimized after a real-face image is input. RCC makes $cos(\pmb{f},\pmb{g}_i)$ higher and $cos(\pmb{f},\pmb{h}_i)$ lower. VAR compels all $cos(\pmb{f},\pmb{g}_i)$ to be equal. DDC urges $cos(\pmb{h}_i,\pmb{g}_j)$ lower and $cos(\pmb{g}_i,\pmb{g}_j)$ higher. }
    \label{fig:main_framework}
\end{figure*}

\textbf{Bayesian Deep Learning.} Bayesian deep learning is one of the commonly used uncertainty estimation methods, which can capture both epistemic uncertainty and aleatoric uncertainty. The network combined with Bayesian deep learning is referred to Bayesian Neural Network (BNN) \cite{denker1990transforming,mackay1992practical}. Later, several approximation approaches, such as variational inference and MC dropout, were presented to model the uncertainty \cite{graves2011practical,blundell2015weight,hernandez2016black,gal2016dropout}.  Recently, methods incorporated with Bayesian deep learning have been applied to various fields, such as recommender systems \cite{li2017collaborative}, object tracking \cite{kosiorek2018sequential}, health care \cite{wang2019bidirectional} and salient object detection \cite{tang2021disentangled}. Whereas, there are rare applications of Bayesian deep learning to face anti-spoofing.

\section{Proposed Method}

As shown in Fig. \ref{fig:main_framework}, the \textbf{Feature Generation and Hypothesis Verification} framework contains a traditional Feature Extraction Backbone, two novel Feature Generation Networks and two powerful hypothesis verification modules: 1) \textbf{Feature Hypothesis Verification Module} estimates the possibility that the input face comes from the real-face feature space. 2) \textbf{Gaussian Hypothesis Verification Module} directly measures the KL divergence between the distribution of the input face and that of real faces in the latent space.

\subsection{Feature Extraction Backbone}

In our framework, the feature extraction backbone takes a single face image $I\in [0,255]^{3\times H \times W}$ as input and outputs the liveness feature vector $\pmb{f}\in\mathbb{R}^{C_f}$, where $H\times W$ is the spatial size and $C_f$ is the dimension of the feature vector.

In the training period, many existing works prefer to directly constrain this feature vector $\pmb{f}$ to obtain a shared feature space. 
However, after considering the diversity of multiple domains, we still find it is too difficult to construct a perfectly generalized feature mapping via limited training data. After all, the raw image space $[0,255]^{3\times H \times W}$ is so large that we can only get a fraction of samples for training.

Instead, we agree only real faces in multiple domains are similar, so we merely attempt to classify real and non-real faces. To achieve it, we mainly construct the real-face distribution and secondarily make the known-attack distribution the auxiliary by means of feature generation networks.

\subsection{Feature Generation Network}

Inspired by GAN \cite{goodfellow2014generative}, 
we design two generation networks to fit the distribution of real faces and known attacks. Nevertheless, considering that noisy or blurry images might be generated \cite{di2021pixel} and the detailed information in images is essential to the FAS task, we urge the two generation networks to directly generate features instead of raw face images.

The two feature generation networks take a latent vector $\pmb{z}\in\mathbb{R}^{C_z}$ as input, where $C_z$ is the dimension of the latent vector and $\pmb{z}\sim \mathcal{N}(\pmb{0},\pmb{I})$ is sampled from the standard multivariate normal distribution. One feature generation network generates a real-face feature vector $\pmb{g}\in\mathbb{R}^{C_f}$ and the other generates a known-attack feature vector $\pmb{h}\in\mathbb{R}^{C_f}$.

For as much as these generated real-face and known-attack features are not extracted from real collected images, it is more appropriate to name the generated features as hypotheses. For the sake of effective hypotheses, the feature generation networks are constrained in regard to both feature space and feature distribution with the assist of the following two hypothesis verification modules.

\subsection{Feature Hypothesis Verification Module}
\label{sec:FHVM}

In an attempt to optimize the feature extraction backbone and the feature generative network in terms of feature space, Feature Hypothesis Verification Module is proposed, which evaluates the consistency of correlations between each real-face hypothesis and the input face feature to determine whether the input face comes from real-face feature space. In our definition, the real-face feature space is a space composed of features which have high consistency of correlations with real-face hypotheses.

Intuitively, the correlations between real-face features should be high, while the correlations between real-face features and non-real-face features should be low. In fact, after obtaining the face feature $\pmb{f}$ and the real-face hypothesis $\pmb{g}$, we utilize cosine similarity to measure the correlation: 

\begin{equation}
\label{ori_cos}
    cos(\pmb{f}, \pmb{g}) = \frac{\pmb{f}\cdot \pmb{g}}{\left\| \pmb{f}\right\|_2 \cdot \left\| \pmb{g}\right\|_2}.
\end{equation}

In pursuit of robust consistency estimation, we take multiple hypotheses into account. And then there will be three situations: (1) If the input face feature has high correlations with each real-face hypothesis, the input face will be thought to be in the same feature space as the real faces. (2) If the input face feature has low correlations with each real-face hypothesis, the input face will also be regarded as a sample from the real-face space but not from the real-face distribution. This situation will be discussed in Sec. \ref{sec:GHVM}. (3) If some correlations are high and the others are low, it will mean that the input face feature space and the real-face feature space have some intersecting subspaces. For this situation, we can determine how these two feature spaces match via measuring the consistency of correlations.

Concretely, we simultaneously sample $N$ Gaussian vectors $\{\pmb{z}_1,\pmb{z}_2,\cdots,\pmb{z}_N\}$ to generate $N$ real-face hypotheses $\{\pmb{g}_1,\pmb{g}_2,\cdots,\pmb{g}_N\}$ and calculate $N$ cosine similarities between $\pmb{f}$ and each hypothesis. By reason that mathematical variance can capture the consistency, we make the variance of cosine similarities the indicator of space intersection size as Eq. \ref{ori_var}. For real-face inputs, the variance should be small. As shown in Fig. \ref{fig:main_framework}, the included angles between the input face feature and each real-face hypothesis tend to be similar. On the contrary, for non-real-face inputs, the variance is usually large and those included angles tend to be different.

\begin{equation}
\label{ori_var}
    VAR = \frac{1}{N-1}\sum_{i=1}^{N}(cos(\pmb{f},\pmb{g}_i)-\frac{\sum_{j=1}^{N}cos(\pmb{f},\pmb{g}_j)}{N})^2
\end{equation}

Even though the variance constraint can not distinguish the first and the second situations above, it has ability to identify the third situation which is the most common situation in actual usage. Moreover, we will partition the first and the second situations in Sec. \ref{sec:GHVM}. It should be noted that the variance constraint does not apply to known-attack hypotheses, because there are various attack types and any attack is not necessarily highly correlated with other known attacks.




Finally, we conclude the advantages of the FHVM by revisiting some domain generalization methods \cite{jia2020single,yang2021few}. These methods map real faces and known attacks to a shared space, and agree that the scores of real faces should be close to 1 and those of attacks should be close to 0. But in this way, if any input face from unknown domains is not mapped to the known region in the shared space, the model will probably output an inaccurate score. Instead, the FHVM urges the feature extraction backbone to map real faces and non-real faces to different spaces. Therefore, before the final prediction is given, we can utilize the variance to check whether the input face belongs to the real-face feature space, which brings on better reliability.

\subsection{Gaussian Hypothesis Verification Module}
\label{sec:GHVM}


Since the distribution of real faces is a distribution composed of features which have high correlations with real-face hypotheses and is only a manifold in the real-face space, not all hypotheses in the real-face space are right real-face features, which explains the second situation illustrated in Sec. \ref{sec:FHVM}. 
To alleviate this issue, we introduce Gaussian Hypothesis Verification Module which brings in two constraints and measures KL divergence between the distribution of input faces and that of real faces.

\textbf{Relative Correlation Constraint.} Considering that the real-face distribution is only a part of meaningful regions in the real-face space, we take the attitude that it is necessary to increase constraints to make this distribution more accurate. Intuitively, for real-face inputs, the correlations with real-face hypotheses should be higher than those with known-attack hypotheses. As for non-real-face inputs, the goal is the opposite. Thus, we propose Relative Correlation Constraint (RCC) in a cross-entropy-like form. After formula derivation, the constraint is represented by Eq. \ref{eq_iter}, where $y'$ is 1 for real-face inputs and is 0 for non-real-face inputs. In particular, Fig. \ref{fig:main_framework} gives an example of how two types of hypotheses are optimized with respect to the given real-face input.

\begin{equation}
\label{eq_iter}
\begin{split}
    RCC = \frac{1}{N}\sum_{i=1}^{N}(ln(e^{cos(\pmb{f},\pmb{g}_i)}+e^{cos(\pmb{f},\pmb{h}_i)})\\
    -y' cos(\pmb{f},\pmb{g}_i)-(1-y')cos(\pmb{f},\pmb{h}_i))
\end{split}
\end{equation}

\textbf{Distribution Distance Measurement.} We calculate KL divergence between the input face distribution and the standard normal distribution in the latent space where $\pmb{z}$ is sampled. For real-face inputs, the $RCC$ losses are usually small, so the modifications of $\pmb{z}_i$ are minor. Considering $\pmb{z}^{(0)}$ are sampled from the standard normal distribution, $\pmb{z}^{(M)}$ would also obey the standard normal distribution. On the contrary, for non-real-face inputs, the modifications are so major that $\pmb{z}^{(M)}$ would obey another unknown distribution.

Enlightened by \citet{schlegl2017unsupervised}, we utilize the Gradient Descent approach to search for the corresponding latent vector of the input face. Specifically, we assume that the input face is a real face, i.e., $y'=1$, and then decrease the RCC loss via optimizing the latent vector $\pmb{z}$ in the hope of acquiring the real-face hypotheses with higher correlations and the known-attack hypotheses with lower correlations. In practice, given an original latent vector $\pmb{z}^{(0)}\sim \mathcal{N}(\pmb{0},\pmb{I})$, the RCC loss can be calculated and a new latent vector $\pmb{z}^{(1)}$ will be obtained via Eq. \ref{z_opt}, where $\alpha$ is the step length for Gradient Descent. The latent vector $\pmb{z}^{(1)}$ is closer to the corresponding latent vector of the input face than $\pmb{z}^{(0)}$. And after $M$ iterations, $\pmb{z}^{(M)}$ is basically able to represent the corresponding latent vector of the input face. Moreover, since $N$ latent vectors are sampled before, $N$ corresponding latent vectors of the input face $\{\pmb{z}_1^{(M)},\pmb{z}_2^{(M)},\cdots,\pmb{z}_N^{(M)}\}$ are obtained.

\begin{equation}
\label{z_opt}
    \pmb{z}^{(1)}=\pmb{z}^{(0)}-\alpha\frac{\partial RCC_{y'=1}^{(0)}}{\partial \pmb{z}^{(0)}}
\end{equation}

For simplification, we assume the corresponding latent vectors of the input face obey a multivariate normal distribution.
KL divergence with the standard normal distribution $\mathcal{N}(0,1)$ can be computed as Eq. \ref{kl_ori}, where $\mu$ and $\sigma^2$ are respectively the estimated mean and variance for a certain dimension in these latent vectors.
After calculating the KL divergences for all dimensions, we average them to acquire the final KL divergence $KL$. We use the difference of two KL divergences $\Delta KL=KL^{(M)}-KL^{(0)}$ to judge whether the input face is real, where $KL^{(0)}$ and $KL^{(M)}$ are the KL divergence before the initial iteration and that after the final iteration, respectively. Note that, $KL^{(0)}$ should be zero theoretically. But allowing for the limited number of samples from the standard normal distribution, it is actually a quite small value. Fortunately, in the experiments, we find that $\Delta KL$ is usually a small value for real-face inputs, while it is a large value for non-real-face inputs.

\begin{equation}
\label{kl_ori}
   KL=-log{\sigma}+\frac{\sigma^2+\mu^2}{2}-\frac{1}{2}
\end{equation}



\textbf{Distribution Discrimination Constraint.} For further improving the discriminative ability of the real-face distribution, we not only make real-face hypotheses more concentrated, but also make the distances between real-face hypotheses and known-attack hypotheses farther. Thus, we impose the distribution discrimination constraint as Eq. \ref{eq_ddc}.

\begin{equation}
\label{eq_ddc}
DDC = \frac{1}{N^2}\sum_{i=1}^{N}\sum_{j=1}^{N}(cos(\pmb{g}_i,\pmb{h}_j)-cos(\pmb{g}_i,\pmb{g}_j))
\end{equation}

\textbf{Overall Loss.} The overall loss function Eq. \ref{eq_overall_loss} is the combination of the variance constraint, the relative correlation constraint and the distribution discrimination constraint, where $\lambda_1$ and $\lambda_2$ are the weights for balance.

\begin{equation}
\label{eq_overall_loss}
\mathcal{L}_{overall} = (2y'-1)\cdot VAR + \lambda_1\cdot RCC + \lambda_2 \cdot DDC
\end{equation}

\subsection{Relation to Bayesian Uncertainty Estimation}

Bayesian deep learning obtains epistemic uncertainty and aleatoric uncertainty by formulating probability distributions over the model parameters and outputs. Especially, the epistemic uncertainty is formulated by placing a prior distribution over the model parameters. Given some inputs, the epistemic uncertainty can be estimated by measuring how much the output varies with these sampled parameters. In previous works \cite{shridhar2018uncertainty, shridhar2018bayesian,farquhar2020radial}, the prior distribution is usually set as a multivariate normal distribution with learnable means and variances. As the model parameters are sampled from the learnt distribution, the variance of the outputs is leveraged to approximate the epistemic uncertainty as Eq. \ref{eu_prior}, where $\pmb{\hat{y}}_t=F^{\hat{W}_t}(\pmb{x})$ is the $t$-th sampled output for random weights $\hat{W}_t\sim q(W)$, and $q(W)$ is the learnt normal distribution.

\begin{equation}
\label{eu_prior}
Epi(\pmb{y})\approx \frac{1}{T}\sum_{t=1}^{T}\pmb{\hat{y}}_t^2-(\frac{1}{T}\sum_{t=1}^{T}\pmb{\hat{y}}_t)^2
\end{equation}

Coincidentally, our proposed approach quite matches the idea of epistemic uncertainty estimation. The feature generation network is equivalent to the parameter generation network for the fully-connected layers, which generates random weights $\hat{W}_t$. And the cosine similarity can be regarded as the sampled output $\pmb{\hat{y}}_t$.

Nevertheless, compared with the previous uncertainty estimation methods which set $q(W)$ to learnable normal distributions, we make $q(W)$ an arbitrary joint distribution which is directly constructed by a generation network. By removing the strong prior assumption that each model parameter obeys an independent normal distribution, we achieve more accurate epistemic uncertainty estimation.

\section{Experiments}

\subsection{Evaluation Basis}

\textbf{Datasets.} Above all, we conduct the cross-dataset testing on four public datasets, i.e., OULU-NPU (denoted as O) \cite{boulkenafet2017oulu}, CASIA-MFSD (denoted as C) \cite{zhang2012face}, Idiap Replay-Attack (denoted as I) \cite{chingovska2012effectiveness} and MSU-MFSD (denoted as M) \cite{wen2015face}. After that, the cross-type testing is carried out on the rich-type dataset, i.e., SiW-M \cite{liu2019deep}.

\noindent \textbf{Classification Bases.}
There are three bases for classification: (1) The mean of $N$ \textit{softmax} outputs derived from $cos(\pmb{f},\pmb{g}_i)$ and $cos(\pmb{f},\pmb{h}_i)$. (2) The variance calculated by Eq. \ref{ori_var}. (3) The $\Delta KL$ claimed in Sec. \ref{sec:GHVM}. In cross-dataset testing, we only use the first score for fair comparison. In cross-type testing, we use all three scores to show the effectiveness of our approach.

\noindent \textbf{Implementation Details.} 
All experiments are conducted via PyTorch on a 32GB Tesla-V100 GPU. For fair comparison, the architecture of our feature extraction network is DepthNet \cite{liu2018learning} which is the same as most alternative methods. The structures of two feature generation networks are the same, each of which consists of two fully-connected layers and a leaky ReLU activation function. The input is only an RGB image (resized to 3x256x256) and has no need of HSV image. During the training period, the framework is trained with SGD optimizer where the momentum is 0.9 and the weight decay is 5e-4. The learning rate is initially 1e-3 and drops to 1e-4 after 50 epochs. The hyper-parameters $\lambda_1$ and $\lambda_2$ are both set to 1. Our source code is available at \url{https://github.com/lustoo/FGHV}.

\subsection{Comparison to Alternative Approaches}


\begin{table*}[t!]
    \caption{Comparison to face anti-spoofing methods on the \textbf{cross-dataset testing} task for domain generalization.}
    \begin{center}
    \scalebox{0.95}{
    \begin{tabular}{ccccccccc}
    \toprule
    \multirow{2}{*}{\textbf{Method}}&
    \multicolumn{2}{c}{\textbf{O\&C\&M to I}}&\multicolumn{2}{c}{\textbf{O\&C\&I to M}}&\multicolumn{2}{c}{\textbf{O\&M\&I to C}}&\multicolumn{2}{c}{\textbf{I\&C\&M to O}}\cr
    \cmidrule(lr){2-3} \cmidrule(lr){4-5} \cmidrule(lr){6-7} \cmidrule(lr){8-9}
    &Hter(\%)&AUC(\%)&Hter(\%)&AUC(\%)&Hter(\%)&AUC(\%)&Hter(\%)&AUC(\%)\cr
    \midrule
    MS\_LBP \cite{LBP03} &50.30&51.64&29.76&78.50&54.28&44.98&50.29&49.31\cr
    Auxiliary(Depth) \cite{liu2018learning} &29.14&71.69&22.72&85.88&33.52&73.15&30.17&77.61 \cr
    MMD-AAE \cite{li2018domain} &31.58&75.18&27.08&83.19&44.59&58.29&40.98&63.08 \cr
    MADDG \cite{shao2019multi} &22.19&84.99&17.69&88.06&24.50&84.51&27.98&80.02 \cr
    SSDG-M \cite{jia2020single} &18.21&94.61&16.67&90.47&23.11&85.45&25.17&81.83\cr
    RFM \cite{shao2020regularized} &17.30&90.48&13.89&93.98&20.27&88.16&16.45&91.16 \cr
    NAS-FAS \cite{yu2020fas} &\textbf{11.63}&\textbf{96.98}&16.85&90.42&15.21&92.64&\textbf{13.16}&\textbf{94.18} \cr
    DRDG \cite{liu2021dual} &15.56&91.79&12.43&95.81&19.05&88.79&15.63&91.75\cr
    D$^2$AM \cite{chen2021generalizable} &15.43&91.22&12.70&95.66&20.98&85.58&15.27&90.87 \cr
    Self-DA \cite{wang2021self} &15.60&90.10&15.40&91.80&24.50&84.40&23.10&84.30 \cr
    ANRL \cite{liu2021adaptive} &16.03&91.04&10.83&96.75&17.85&89.26&15.67&91.90 \cr
    \midrule
    \textbf{Ours}&16.29&90.11&\textbf{9.17}&\textbf{96.92}&\textbf{12.47}&\textbf{93.47}&13.58&93.55\cr
    \bottomrule
    \end{tabular}}
    \end{center}
    \label{tab:four_cross_domain}
\end{table*}

\begin{table*}[htbp]
  \centering
  \caption{Comparison to face anti-spoofing methods on the \textbf{cross-type testing} task for domain generalization.}
     \resizebox{\textwidth}{!}{
    \begin{tabular}{cccccccccccccccc}
    \toprule
    \multirow{2}[4]{*}{\textbf{Method}} & \multirow{2}[4]{*}{\textbf{Metrics(\%)}} & \multirow{2}[4]{*}{Replay} & \multirow{2}[4]{*}{Print} & \multicolumn{5}{c}{Mask Attacks}      & \multicolumn{3}{c}{Makeup Attacks} & \multicolumn{3}{c}{Partial Attacks} & \multirow{2}[4]{*}{\textbf{Average}} \\
\cmidrule(lr){5-9} \cmidrule(lr){10-12} \cmidrule(lr){13-15}          &       &       &       & Half  & Silicone & Trans. & Paper & Manne. & Obfusc. & Imperson. & Cosmetic & Funny Eye & Glasses & Partial &  \\
    \midrule
    \multirow{2}[1]{*}{\shortstack{Auxiliary \\ \cite{liu2018learning}}} & ACER  & 16.8  & 6.9   & 19.3  & 14.9  & 52.1  & 8.0   & 12.8  & 55.8  & 13.7  & 11.7  & 49.0  & 40.5  & 5.3   & 23.6±18.5 \\
          & EER   & 14.0  & 4.3   & 11.6  & 12.4  & 24.6  & 7.8   & 10.0  & 72.3  & 10.1  & 9.4   & 21.4  & 18.6  & 4.0   & 17.0±17.7 \\
    \midrule
    \multirow{2}[2]{*}{\shortstack{DTN \\ \cite{liu2019deep}}} & ACER  & 9.8   & \textbf{6.0}   & 15.0  & 18.7  & 36.0  & 4.5   & 7.7   & 48.1  & 11.4  & 14.2  & 19.3  & 19.8  & 8.5   & 16.8±11.1 \\
          & EER   & 10.0  & \textbf{2.1}   & 14.4  & 18.6  & 26.5  & 5.7   & 9.6   & 50.2  & 10.1  & 13.2  & 19.8  & 20.5  & 8.8   & 16.1±12.2 \\
    \midrule
    \multirow{2}[2]{*}{\shortstack{SpoofTrace \\ \cite{liu2020disentangling}}} & ACER  & \textbf{7.8}   & 7.3   & 7.1   & 12.9  & \textbf{13.9}  & 4.3   & 6.7   & 53.2  & 4.6   & 19.5  & 20.7  & 21.0  & 5.6   & 14.2±13.2 \\
          & EER   & \textbf{7.6}   & 3.8   & 8.4   & 13.8  & 14.5  & 5.3   & 4.4   & 35.4  & \textbf{0.0}   & 19.3  & 21.0  & 20.8  & 1.6   & 12.0±10.0 \\
    \midrule
    \multirow{2}[2]{*}{\shortstack{NAS-FAS \\ \cite{yu2020fas}}} & ACER  & 9.3  & 7.9   & 11.4   & 12.1  & 15.8  & \textbf{1.9}   & 2.7   & 28.5  & 0.4   & 15.1  & \textbf{16.5}  & 16.0  & 3.8   & 10.9±7.8 \\
          & EER   & 9.3  & 6.8   & 9.7   & 11.1  & \textbf{12.5}  & \textbf{2.7}   & \textbf{0.0}   & 26.1  & \textbf{0.0}   & 15.0  & \textbf{15.1}  & 13.4  & 2.3   & 9.5±7.4 \\
    \midrule
    \multirow{2}[2]{*}{\shortstack{DC-CDN \\ \cite{yu2021dual}}} & ACER  & 12.1  & 9.7   & 14.1  & \textbf{7.2}   & 14.8  & 4.5   & \textbf{1.6}   & 40.1  & \textbf{0.4}   & 11.4  & 20.1  & 16.1  & \textbf{2.9}   & 11.9±10.3 \\
          & EER   & 10.3  & 8.7   & 11.1  & \textbf{7.4}   & \textbf{12.5}  & 5.9   & \textbf{0.0}   & 39.1  & \textbf{0.0}   & 12.0  & 18.9  & 13.5  & \textbf{1.2}   & 10.8±10.1 \\
    \midrule
    \multirow{2}[2]{*}{\textbf{Ours}} & ACER  &  8.4 &  7.3  & \textbf{5.2} & 9.8 & 14.2  & 3.2  &   4.1   &  \textbf{16.7} &  1.9& \textbf{9.0} & 18.2 &   \textbf{8.3}    &4.4  & \textbf{8.5±5.1} \\
          & EER  &   9.0 & 8.0  &  \textbf{5.9}   &  9.9  &  14.3   & 3.7  & 4.8   & \textbf{19.3}  & 2.0 &  \textbf{9.2} &  18.9  & \textbf{8.5}  &  4.7 & \textbf{9.1±5.4} \\
    \bottomrule
    \end{tabular}%
    } 
  \label{tab:siwm}%
\end{table*}%

\begin{table}[t!]
\begin{center}
    \caption{Comparison to uncertainty estimation methods.}
    \scalebox{0.95}{
    
    \begin{tabular}{ccc}
    \toprule
    \textbf{Method} & HTER(\%) & AUC(\%)  \\
    \midrule
    Bayesian Neural Network & 20.00  & 88.67  \\
    MC Dropout & 15.00  & 92.50  \\
    Deep Ensemble & \textbf{5.83}  & 95.75  \\
    \midrule
    \textbf{Ours} & 9.17  & \textbf{96.92}  \\
    \bottomrule
    \end{tabular}}
    \label{tab:uncertain_estimate}
\end{center}
\end{table}

\noindent \textbf{Cross-dataset testing.} We strictly follow the previous methods \cite{jia2020single,qin2020learning} and select one dataset for testing and the other three datasets for training. 
As for evaluation, we follow the popular metrics, i.e., the Half Total Error Rate (HTER) and the Area Under Curve (AUC), and let the \textit{Softmax} scores derived from real-face correlations and known-attack correlations become the only basis for classification. Compared to the alternative methods in Table \ref{tab:four_cross_domain}, our FGHV framework has overwhelming performance improvement in two cross-dataset settings and comparable accuracy in the left two settings. Although NAS-FAS \cite{yu2020fas} got a substantial increase in effect by searching a much stronger backbone than DepthNet, 
our framework is able to enhance the effects of cross-dataset scenarios only with the aid of the three constraints, which indicates better generalization. 

It is worth noting that some ideas between our framework and SSDG \cite{jia2020single} are quite similar. Firstly, the asymmetric triplet loss in SSDG and our DDC loss are both aimed at optimizing distances among features. In fact, SSDG only constrained input-face features whose quantity is restricted by the size of datasets. Instead, we choose to restrain considerable generated hypotheses so that the distribution of real faces and that of non-real faces will be accurately constructed and separated. Secondly, the feature generator in SSDG is actually a feature extractor and the discriminator makes extracted features domain-agnostic, while our feature generation networks are real generators that are optimized by FHVM and GHVM.

\noindent \textbf{Cross-type testing.} Strictly following the cross-type testing protocol (13 attacks leave-one-out) on SiW-M, we select out one attack type as the unknown testing type and treat the others as the known training types for each experiment. As for performance metrics, Average Classification Error Rate (ACER) and Equal Error Rate (EER) are utilized. Since ACER describes the practical performance under predeterminate thresholds, we additionally take advantage of the variance and $\Delta KL$ to assist classification. As shown in Table. \ref{tab:siwm}, our framework can significantly improve the overall performance by 2.4\% for ACER and 0.4\% for EER. Meanwhile, the results are more stable. Surprisingly, our framework achieves a great promotion in defense for difficult obfuscation makeup attacks. Furthermore, it is worth noting that the variance and $\Delta KL$ indeed contribute to detecting attacks (ACERs are overall lower than EERs) so they are quite practical in actual usage. Consequently, excellent reliability is guaranteed.

\noindent \textbf{Comparison to uncertainty estimation methods.} Since the FHVM plays a similar role in estimating epistemic uncertainty, we compare our framework with three representative uncertainty estimation methods, i.e., Bayesian Neural Network \cite{kendall2017uncertainties}, MC Dropout \cite{gal2016dropout} and Deep Ensemble \cite{lakshminarayanan2017simple}. These methods are reproduced on O\&C\&I to M setting and the comparison results are shown in Table \ref{tab:uncertain_estimate}. It has been proved that the strong prior assumption (i.e., normal distribution) in Bayesian Neural Network limits the ability of models. As for MC Dropout, it is too simple to bring exciting improvement. Moreover, Deep Ensemble outperforms in HTER but underperforms in AUC, which means that our method and Deep Ensemble are comparable in terms of overall performance. However, Deep Ensemble constructs the distribution for the whole convolution and fully-connected layers, which explains why it is more time-consuming and memory-consuming.

\subsection{Ablation Study}



\begin{table}[t!]
\begin{center}
    \caption{Comparison of different constraints.}
    \scalebox{0.83}{ 
    \begin{tabular}{ccccc}
    \toprule
    RCC&VAR&DDC & HTER(\%) & AUC(\%)  \\
    \midrule
    \checkmark&&&14.16&91.12   \\
    &\checkmark&&33.33&64.50  \\
    &&\checkmark&19.12&86.83   \\
    \checkmark&\checkmark&&10.00&93.92  \\
    \checkmark&&\checkmark&10.83&93.50  \\
    &\checkmark&\checkmark&12.50&94.23  \\
    \checkmark&\checkmark&\checkmark&\textbf{9.17}&\textbf{96.92} \\
    
    \bottomrule
    \end{tabular}}

    \label{tab:each_constraint}
\end{center}
\end{table}

\begin{figure}[t!]
    \centering
    \includegraphics[width = \linewidth]{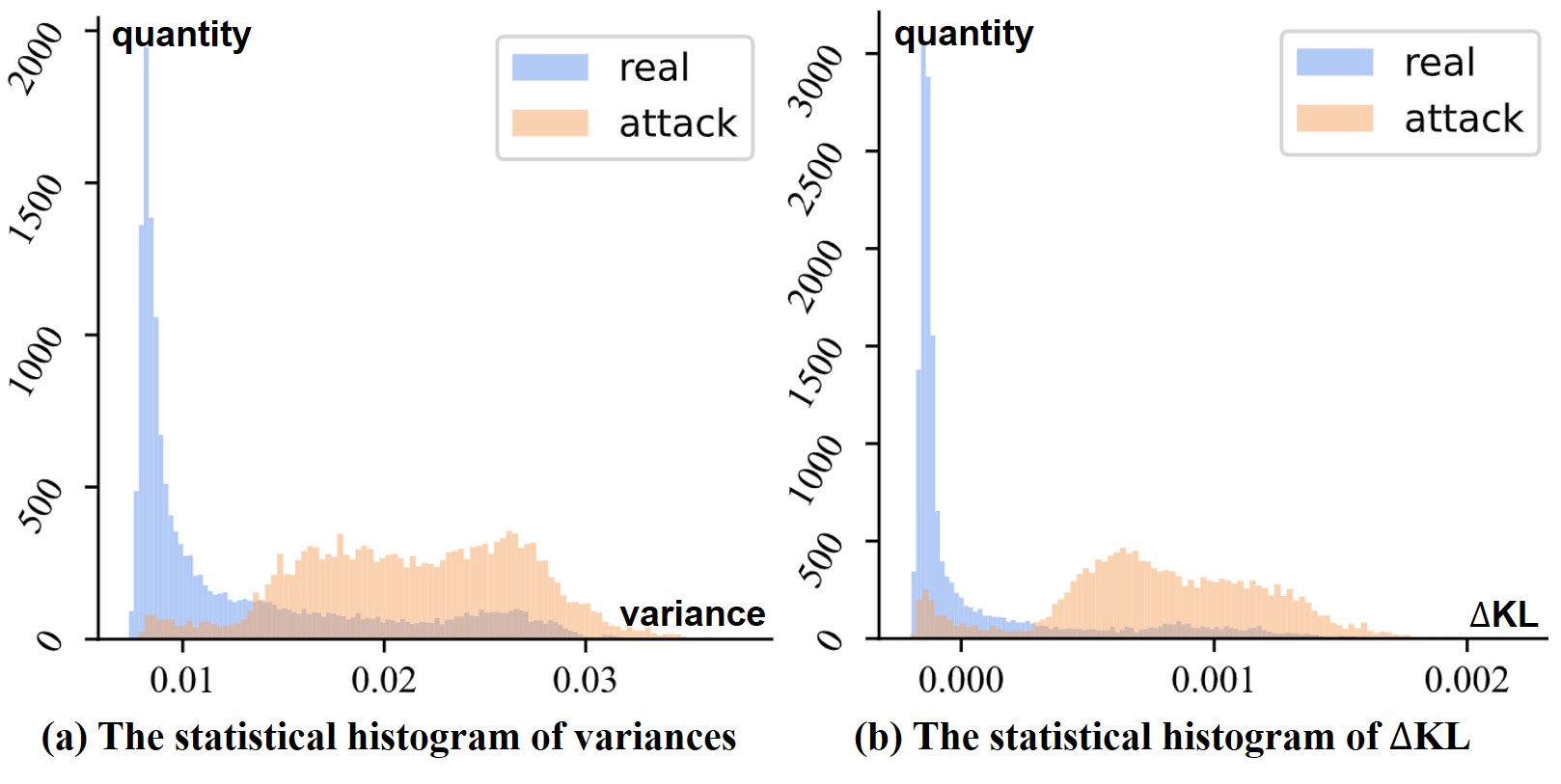} 
    \caption{The quantity varies with variance and $\Delta KL$.}
    \label{fig:two_hvm}
\end{figure}

\textbf{What is the effect of each constraint?}\quad In an effort to explore the effects of three constraints in our framework, we do ablation studies on O\&C\&I to M setting. For each experiment, we select different constraints to optimize the same framework and then use HTER and AUC for evaluation. As indicated in Table \ref{tab:each_constraint}, the framework with single RCC constraint can be regarded as the baseline because the RCC is similar with cross entropy loss. The result of single VAR constraint is unsatisfactory on account of not constructing real-face distributions. But if the VAR constraint is accompanied with the RCC, the effect will be improved a lot. Additionally, the usage of the DDC makes the real-face distribution more discriminative so that the final effect reaches the current best. In summary, the three constraints are indispensable and boost the effectiveness from different aspects.

\noindent \textbf{Is the FHVM discriminative?}\quad By reason that the FHVM utilizes the consistency of correlations to estimate to what extent the input face belongs to the real-face feature space, we analyze how the quantities of real faces and attacks vary with the variance of correlations on I\&C\&M to O setting. As shown in Fig. \ref{fig:two_hvm}(a), the variances of correlations between real-face inputs and real-face hypotheses are much lower than those between attack inputs and real-face hypotheses. The results point out that the uncertainty estimation is indeed carried out and the predictions of real-face inputs are more certain. Hence, the discrimination of the FHVM is proved.


\noindent \textbf{Is the GHVM discriminative?}\quad In an attempt to probe the discrimination of the GHVM, we utilize Gradient Decent strategy to find the corresponding latent vectors of the input face on I\&C\&M to O setting as explained in Sec. \ref{sec:GHVM}. The number of iterations $M$ is 15 and the step length $\alpha$ is 1. After acquiring $\Delta KL=KL^{(15)}-KL^{(0)}$, we create a histogram to reveal how the quantity varies with $\Delta KL$. As shown in Fig. \ref{fig:two_hvm}(b), it is easy to find that $\Delta KL$s of real faces are generally small while $\Delta KL$s of fake faces are usually large, which is consistent with our expectation. And if the threshold is chosen appropriately, the separation of real faces and attacks can be achieved successfully. Therefore, the results demonstrate that the GHVM is discriminative.

\begin{table}[t!]
\begin{center}
    \caption{The impacts of both feature generation networks.}
    \scalebox{0.95}{
    
    \begin{tabular}{ccc}
    \toprule
    \textbf{Method} & HTER(\%) & AUC(\%)  \\
    \midrule
    No FGN & 25.00  & 83.08  \\
    Only RF FGN & 15.83  & 86.67  \\
    Only KA FGN & 20.00  & 90.67  \\
    \midrule
    \textbf{Both FGNs} & \textbf{9.17}  & \textbf{96.92}  \\
    \bottomrule
    \end{tabular}}
    \label{tab:impact_two_fgns}
\end{center}
\end{table}

\begin{figure}[t!]
    \centering
    \includegraphics[width = 0.64\linewidth]{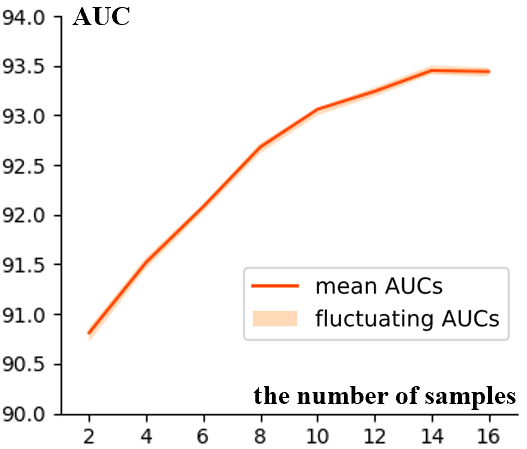} 
    \caption{The impact of the number of samples.}
    \label{fig:sample_latent_vector}
\end{figure}

\noindent \textbf{Is it necessary to introduce two feature generation networks?}\quad Aiming to explore the necessity of both real-face and known-attack feature generation networks, we launch a probe into the impacts of both feature generation networks (FGN) on O\&C\&I to M setting. The structure without any FGN (No FGN) is the same with the conventional binary classification network. The structure equipped with known-attack FGN (Only KA FGN) can utilize RCC constraint, while the structure equipped with real-face FGN (Only RF FGN) can benefit from both RCC and VAR constraints. And all three constraints cannot be used until two FGNs (Both FGNs) are introduced at the same time. The results shown in Table \ref{tab:impact_two_fgns} prove that two feature generation networks accompanied with three constraints can bring the greatest improvement. Thus, the necessity is self-evident.

\noindent \textbf{Can more sampled latent vectors bring better performance or stability?}\quad To answer this question, 
we change the number of sampled latent vectors to repeat training and testing on I\&C\&M to O setting for several times (i.e., twenty times). For performance assessment, we average the twenty AUCs on the testset. For stability assessment, we utilize the maximum and the minimum of the twenty AUCs to reflect the fluctuation of effects. As depicted in Fig. \ref{fig:sample_latent_vector}, with the number of samples increasing, the effects are enhanced a lot. Furthermore, the training process and evaluation results become more stable. Taken time consumption and accuracy into account, the optimal number of samples is 14.
Not only is it possible for us to compromise accuracy and speed, but applications can benefit from the reproducible experiments and the robust models.


\begin{figure}[t!]
    \centering
    \includegraphics[width = \linewidth]{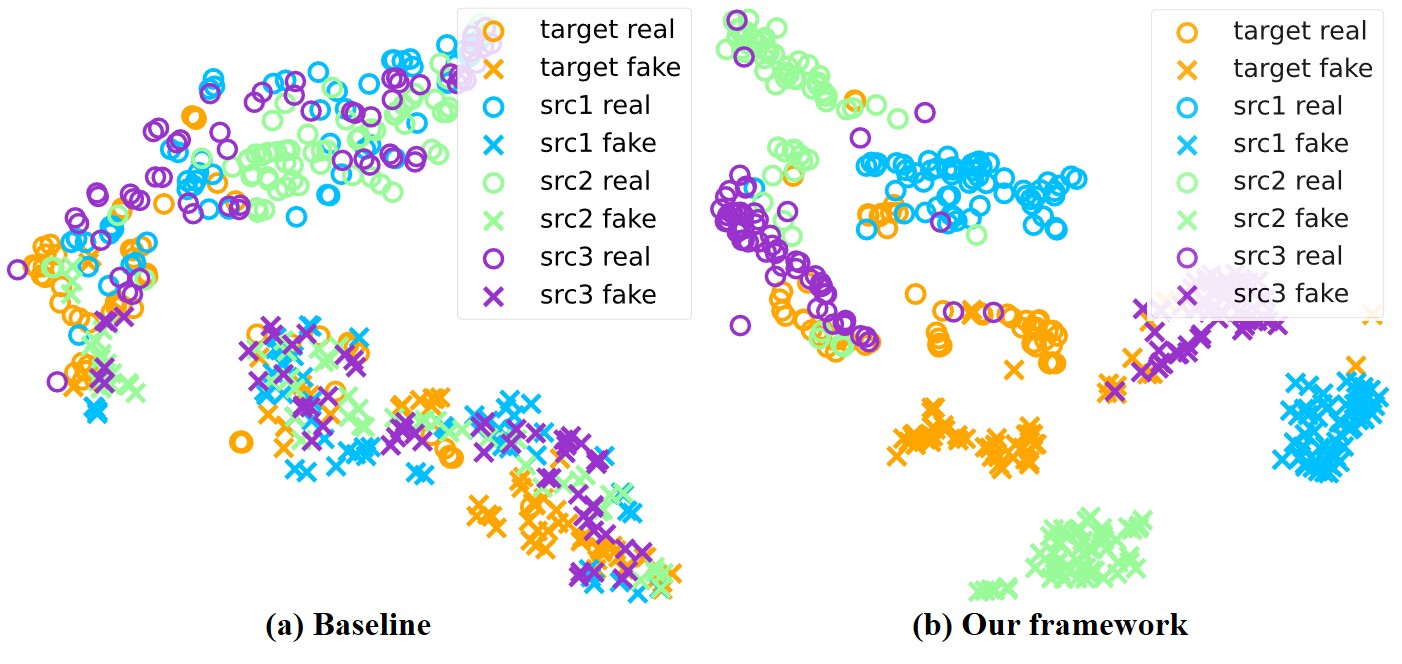} 
    \caption{The t-SNE visualization.}
    \label{fig:exp_tsne}
\end{figure}

\noindent \textbf{Will the feature extraction backbone be improved coincidentally?}\quad Although we are mainly committed to improving the feature generation networks and attach less significance to the feature extraction backbone, we intend to validate whether the feature extraction backbone is also improved coincidentally. To confirm it, we visualize the features of input faces via t-SNE \cite{van2008visualizing}, which is depicted in Fig. \ref{fig:exp_tsne}. We conduct the ablation study on I\&C\&M to O setting and treat the model which is optimized only with cross entropy loss as the baseline method. According to the visualization, the baseline method classifies real faces and attacks well just in the source domains but poorly in the target domain. On the contrary, for our approach, even though attacks are not gathered together, the real-face regions are well constructed, which makes it easier to distinguish real faces and non-real faces. That's to say, our feature extraction backbone is promoted passingly.

\section{Conclusion}

In this paper, we propose a feature generation and hypothesis verification framework for FAS. Firstly, for the purpose of generalization, we regard the FAS task as the classification of real faces and non-real faces. Then, two feature generation networks are devised for the first time and two hypothesis verification modules are designed to estimate to what extent the input face belongs to the feature space and the distribution of real faces. Finally, we analyze the framework from the viewpoint of Bayesian uncertainty estimation and demonstrate the reliability of the framework. Qualitative and quantitative analyses show our framework outperforms the state-of-the-art approaches.

\bibliography{aaai22}

\end{document}